\pdfoutput=1

\documentclass[11pt]{article}

\usepackage{EMNLP2022}

\usepackage{times}
\usepackage{latexsym}

\usepackage{graphicx}
\usepackage{multirow}
\usepackage{amssymb}
\usepackage{amsmath}
\usepackage{multicol}
\usepackage{caption}
\usepackage{subcaption}
\usepackage{algorithm}
\usepackage{CJKutf8}
\usepackage{algpseudocode}
\usepackage{listings}
\usepackage{float}
\usepackage{times}
\usepackage{booktabs}
\usepackage{fancyvrb}
\usepackage{fdsymbol}
\usepackage{float}
\usepackage{xcolor}
\usepackage{pifont}
\newcommand{\cmark}{\ding{51}}%
\newcommand{\xmark}{\ding{55}}%

\usepackage[T1]{fontenc}

\usepackage[utf8]{inputenc}

\usepackage{microtype}

\usepackage{inconsolata}

\usepackage[colorinlistoftodos,prependcaption,textsize=tiny]{todonotes}
\colorlet{punct}{red!60!black}
\definecolor{background}{HTML}{EEEEEE}
\definecolor{delim}{RGB}{20,105,176}
\colorlet{numb}{magenta!60!black}
\definecolor{codegreen}{rgb}{0,0.6,0}
\definecolor{codegray}{rgb}{0.5,0.5,0.5}
\definecolor{codepurple}{rgb}{0.58,0,0.82}
\definecolor{backcolour}{rgb}{0.95,0.95,0.92}
\lstdefinestyle{mystyle}{
    commentstyle=\color{codegreen},
    keywordstyle=\color{magenta},
    numberstyle=\tiny\color{codegray},
    stringstyle=\color{codepurple},
    basicstyle=\ttfamily\tiny,
    breakatwhitespace=false,         
    breaklines=true,                 
    captionpos=b,                    
    keepspaces=true,                 
    numbers=left,                    
    numbersep=5pt,                  
    showspaces=false,                
    showstringspaces=false,
    showtabs=false,                  
    tabsize=2
}

\title{\includegraphics[width=.04\linewidth]{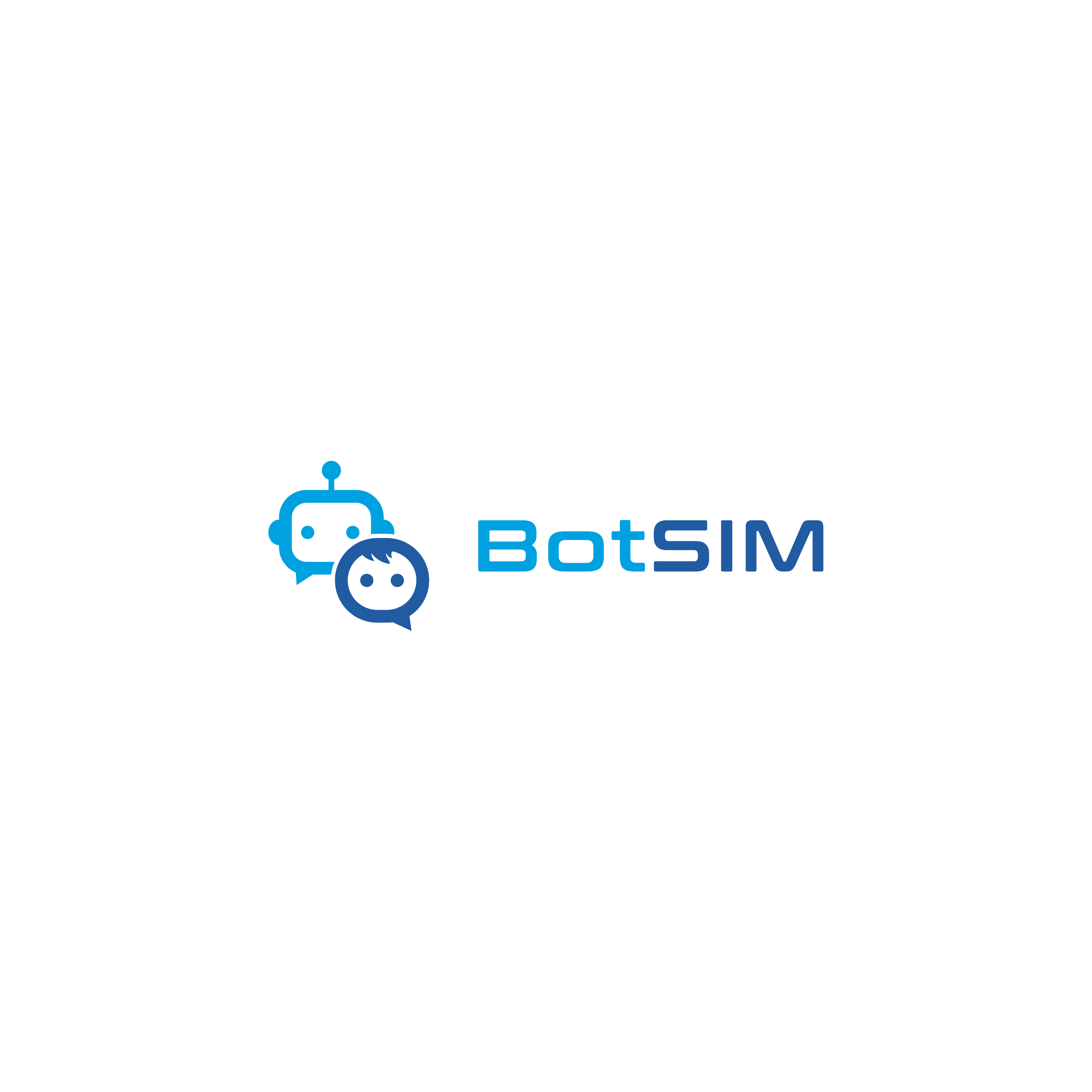}~BotSIM: An End-to-End Bot Simulation Framework\\ for Commercial Task-Oriented Dialog Systems}
 
 \author{Guangsen Wang$^\clubsuit$ \quad Samson Tan$^\Diamond$\thanks{\; Work done at Salesforce Research.}  \quad Shafiq  Joty$^\clubsuit$ \quad Gang Wu$^\clubsuit$ \quad Jimmy Au$^\clubsuit$ \quad Steven Hoi$^\clubsuit$ \\
        $^\clubsuit$Salesforce Research\\
       $^\Diamond$AWS AI Research \& Education\\ \texttt{\{guangsen.wang, sjoty, jimmy.au, gang.wu, shoi\}@salesforce.com}}

\begin{document}
\maketitle
\begin{abstract}
We present BotSIM, a data-efficient end-to-end\textbf{ Bot SIM}ulation toolkit for commercial text-based task-oriented  dialog (TOD) systems. BotSIM consists of three major components: 1) a \textit{Generator} that can infer semantic-level dialog acts and entities from bot definitions and generate user queries via model-based paraphrasing; 2) an agenda-based dialog user \textit{Simulator} (ABUS) to simulate conversations with the dialog agents; 3) a \textit{Remediator} to analyze the simulated conversations, visualize the bot health reports and provide actionable remediation suggestions for bot troubleshooting and improvement.  
We demonstrate BotSIM's effectiveness in end-to-end evaluation, remediation and multi-intent dialog generation via case studies on two commercial bot platforms. BotSIM's ``generation-simulation-remediation'' paradigm accelerates the end-to-end bot evaluation and iteration process by: 1) reducing manual test cases creation efforts; 2) enabling a holistic gauge of the bot in terms of NLU and end-to-end performance via extensive dialog simulation; 3) improving the bot troubleshooting  process with actionable suggestions. 
{A demo of our system can be found at \url{https://tinyurl.com/mryu74cd} and a demo video at \url{https://youtu.be/qLi5iSoly30}. We have open-sourced the toolkit at \url{https://github.com/salesforce/botsim}.}
\end{abstract}

\section{Introduction}

\begin{figure}[!t]
  \centering
\includegraphics[width=7.8cm]{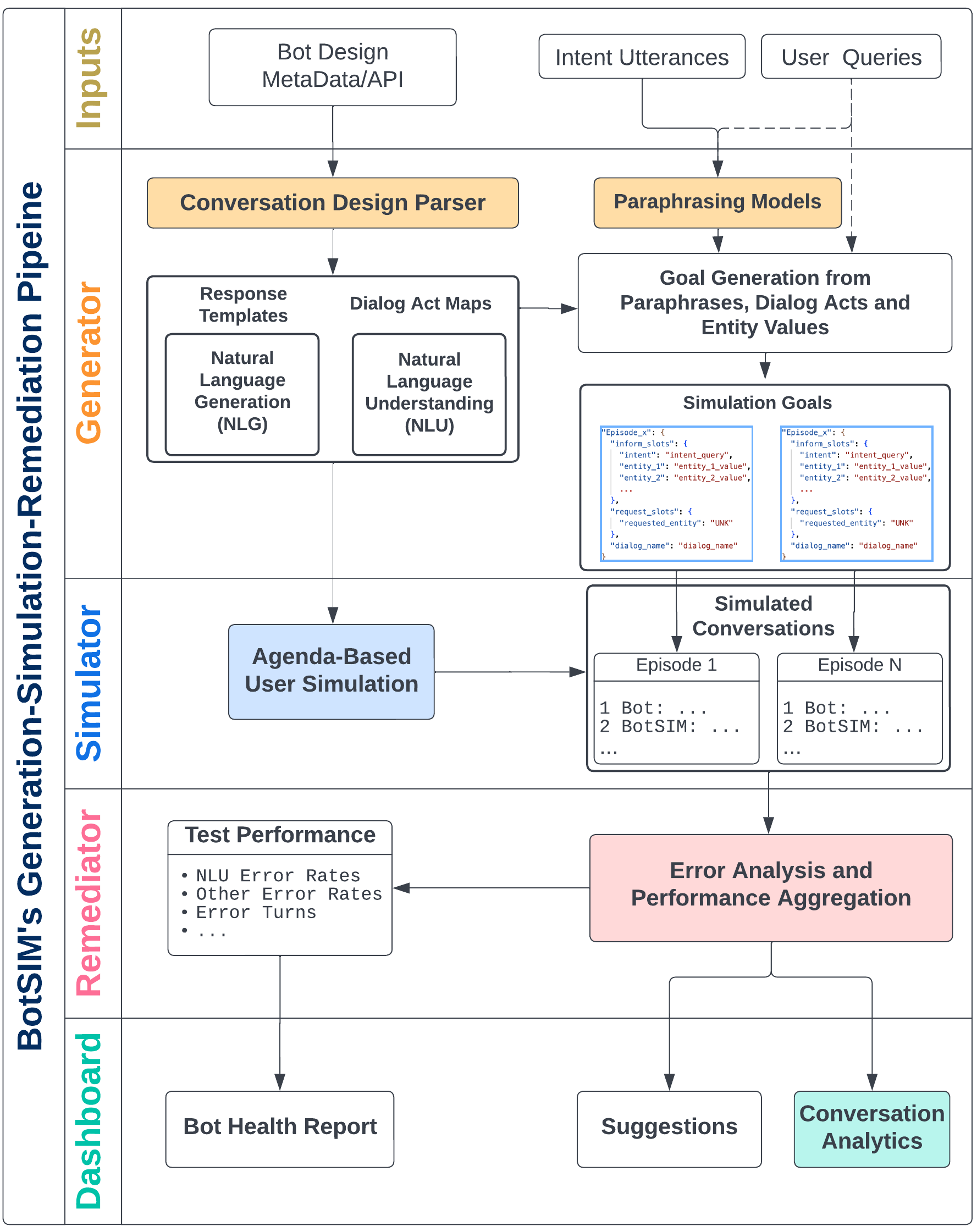} 
  \caption{BotSIM overview including the generator, simulator, and remediator. The dotted (optional) paths from users can be used for bot performance monitoring: they can provide production chat logs or manually crafted utterances when creating evaluation goals. }
  \label{fig:architecture}
  \vspace{-1em}
\end{figure}

The typical dialog system development cycle consists of dialog design, pre-deployment testing, deployment, performance monitoring,  model improvement and iteration. As in any production software system, effective and comprehensive testing at all stages is of paramount importance. 
Unfortunately, \emph{evaluating and troubleshooting} production TOD systems is still a largely manual process requiring large amount of human conversations with the systems.
This process is time-consuming, expensive, and inevitably fails to capture the breadth of language variation present in the real world \citep{tan-etal-2021-reliability}.  The time- and labor-intensive nature of such an approach is further exacerbated when the developer significantly changes the dialog flows, since new sets of test dialogs will need to be created~\citep{IBM-Watson}. Performing comprehensive end-to-end bot evaluation is highly challenging due to the need for additional annotation efforts. Finally, there is a lack of analytical tools for interpreting  test results and troubleshooting  underlying bot issues.

To address these limitations, we present \emph{BotSIM}, a  \textbf{Bot SIM}ulation environment for data-efficient end-to-end  commercial bot evaluation, remediation via multi-intent dialog generation and agenda-based dialog user simulation~\citep{schatzmann-etal-2007-agenda}. 
BotSIM consists of three major modules, namely Generator, Simulator, and Remediator (Figure~\ref{fig:architecture}). We use a pretrained sequence-to-sequence T5 model~\citep{zhang2019pegasus,raffel2020exploring} in the  Generator to simulate lexical and syntactic variations in user queries via paraphrasing. The Generator is also responsible to generate  various templates needed by the Simulator. To make BotSIM more platform- and task- agnostic, we adopt dialog-act level ABUS to simulate conversations with bots via APIs. {The dialog acts are automatically inferred by the Generator via a unified interface to convert bot designs of different platforms to a universal graph representation. The graph has all dialogs as nodes and their transitions as edges. Through graph traversal, BotSIM offers a principled and scalable approach to  generating and exploring  multi-intent conversations. Not only can the conversation path generation greatly increase evaluation coverage for troubleshooting dialog errors caused by faulty designs (\emph{e.g.,} unexpected dialog loops), it is also valuable for bot design improvements.}
The Remediator  summarizes bots' health status in a dashboard  for easy comprehension. It also analyzes   the simulated conversations to identify any issues and further provides actionable suggestions to remedy them.

BotSIM's ``generation-simulation-remediation'' paradigm  can significantly accelerate bot development and evaluation, reducing human efforts, cost and time-to-market.
Our  contributions include:
\begin{itemize} 
    \vspace{-0.5em}
    \setlength{\itemsep}{-3pt}
    \item We propose BotSIM, a modular, data-efficient  bot simulation environment.
    To the best of our knowledge, this is the first work focused on end-to-end evaluation, diagnosis and remediation of commercial bots via ABUS.
    \item {BotSIM offers a principled approach to generating and simulating multi-intent dialogs for comprehensive evaluation coverage and better bot design.}
   \item We finetuned a T5 paraphrasing model on par with the state-of-the-art performance to generate diverse user responses for greater test coverage of language variation.
    \item An easy-to-use Streamlit\footnote{\url{https://streamlit.io/}} Web App with Flask back-end and SQL database is developed  for bot practitioners. The App can be deployed as a docker container or to Heroku\footnote{\url{https://www.heroku.com}}. 
\end{itemize}

\section{Related Work}

There are two main categories of dialog user simulators, namely the agenda-based user simulator (ABUS)~\citep{schatzmann-etal-2007-agenda, LiLDLGC16-TCBot, shi2019build, zhu2020convlab2, liu2021robustness,DBLP:journals/corr/abs-1801-04871} and recent neural-based user simulator (NUS)~\citep{AsriHS16-seq2seq-us, crook2017sequence, NUS, gur2018user,DBLP:journals/corr/abs-1711-10712}. 
Since BotSIM is designed to support commercial bot evaluation and remediation, we focus on the review of the testing capacities offered by some existing bot platforms rather than the simulators. Recently, 
ABUS  is also used in Amazon's Alexa conversation~\cite{DBLP:conf/naacl/AcharyaAAABBCCF21} for training an end-to-end dialog agent, which is also beyond the scope of our discussion.
 

\begin{table*}[!htbp]
\resizebox{\textwidth}{!}{%
\tiny
\begin{tabular}{ccccccccc}
\multirow{2}{*}{} &
  \multicolumn{2}{c}{Methods} &
  \multicolumn{2}{c}{Stages} &
  \multicolumn{2}{c}{Automation} &
  \multicolumn{2}{c}{Metrics} \\ \toprule 
 &
  Regression &
  End-to-end &
  Pre-deployment &
  Monitoring &
  \begin{tabular}[c]{@{}c@{}}Test case \\ curation\end{tabular} &
  \begin{tabular}[c]{@{}c@{}}User \\ Simulation\end{tabular} &
  NLU &
  \begin{tabular}[c]{@{}c@{}}Task \\ Completion\end{tabular} \\ \midrule
CX & \cmark &      &      & \cmark  &      &      &      &      \\
Watson    & \cmark  &      &      & \cmark  &      &      & \cmark  & \cmark  \\
Botium        & \cmark  &      &      &      &      &      & \cmark  &      \\
BotSIM        & \cmark  & \cmark  & \cmark  & \cmark  & \cmark  & \cmark  & \cmark  & \cmark  \\ \bottomrule
\end{tabular}%
}
 \caption{Comparison of bot evaluation capabilities of the reviewed commercial bot platforms}
 \label{tab:platform-cmp}
 \vspace{-1em}
\end{table*}
\subsection{IBM Watson Assistant}
IBM Watson assistant offers a suite of open-source Python libraries and notebooks to help analyze customer bots using manually created or annotated test cases~\citep{IBM-Watson}.
An exemplar test case used for  the standard regression testing is given in  Table~\ref{tab:ibm}. 
\begin{table}[!htbp]
\begin{tabular}{c|c}
\toprule
\textbf{User:}            & May I book a flight to New York?     \\ \midrule
\multirow{2}{*}{\textbf{Labels}} & \textbf{Intent: \#flight}            \\ \cline{2-2} 
& \textbf{Entity: @Destination}        \\ \hline
\textbf{Bot:}             & When would you like to depart? \\ \bottomrule
\end{tabular}
\caption{Example of IBM Watson assistant test case}
\label{tab:ibm}
\vspace{-1em}
\end{table}
Given the annotated conversations, the notebooks offer some analytical functions to compute two metrics, namely coverage (NLU) and effectiveness (task completion) to monitor the bot performance. However, the manual annotation and analysis still require significant expertise and involvement of bot teams. Large scale automatic pre-deployment performance evaluation and analysis are also infeasible since there may not be enough chat logs.

\subsection{Google DialogFlow CX}
\label{sec:cx}
\begin{figure}[t]
  \centering
\includegraphics[width=8cm]{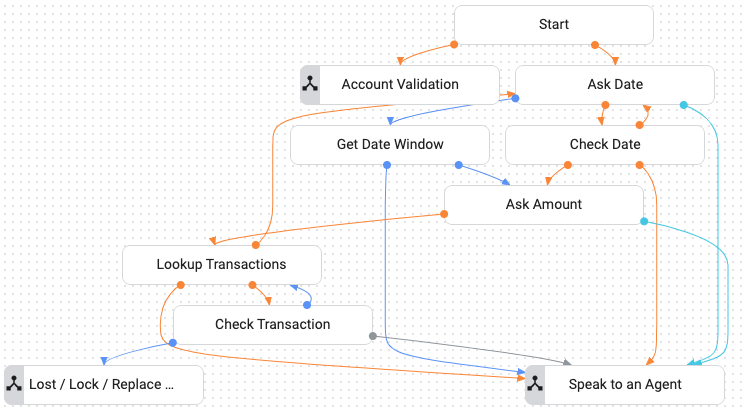} 
  \caption{``Investigate Charges'' flow of the DialogFlow CX pre-built  ``Financial Service Agent'' mega-agent}
  \label{fig:cx-graph}
\end{figure}
CX offers a built-in regression testing environment for users to create test dialogs and perform regression testing. 
To create ``golden'' test cases, users need to manually chat with the bot and annotate each system turn with correct intent, entity and dialog transitions. During regression testing, each bot response is matched against the golden labels to detect regressions.
To achieve good regression testing coverage, users have to ``design'' testing dialogs to cover as many conversation paths as possible.  However, the number of paths grows exponentially with the number of intents and dialog branches, making it almost impossible to craft testing cases for all paths (Figure~\ref{fig:cx-graph}). 
For end-to-end performance evaluation, even more annotated dialogs are needed to cover the language variation in user responses, which will greatly intensify the manual efforts. 

\subsection{Botium: Bots Testing Bots}
Botium\footnote{\url{https://www.botium.ai/}} offers a unified platform for regression testings of various bot platforms via platform-specific connectors and  ``Botium scripts (test cases)'' (Appendix~\ref{test-case}).
The core component, dubbed as ``Botium Box'', analogous to a bot testing ``IDE'', can be used to connect different bot  platforms, create testing cases and conduct regression testings. 
However, the testing capability is constrained by the underlying platforms. For example, users may be still required to design testing dialogs manually. Therefore, Botium  cannot perform large scale end-to-end pre-deployment testing. 
%

The overall comparison of different platforms is given in Table~\ref{tab:platform-cmp}. Most current platforms only focus on regression testing.
While regression testing is important to ensure correct and consistent system behaviours,  it is also vital to perform pre-deployment evaluation to avoid poor user adoption and retention rate. Although some  platforms  are capable of computing turn-level NLU metrics, they require significant manual efforts in curating or annotating test cases. In addition, the NLU metrics do not directly translate to the end-to-end goal completion performance. We will show how BotSIM can help circumvent these limitations via large scale automatic dialog generation and simulation.

\vspace{-0.5em}
\section{BotSIM} 
BotSIM system overview is shown in Figure~\ref{fig:architecture}. 
\vspace{-0.5em}
\subsection{Generator}

\begin{figure}[!t]
  \includegraphics[width=7.8cm]{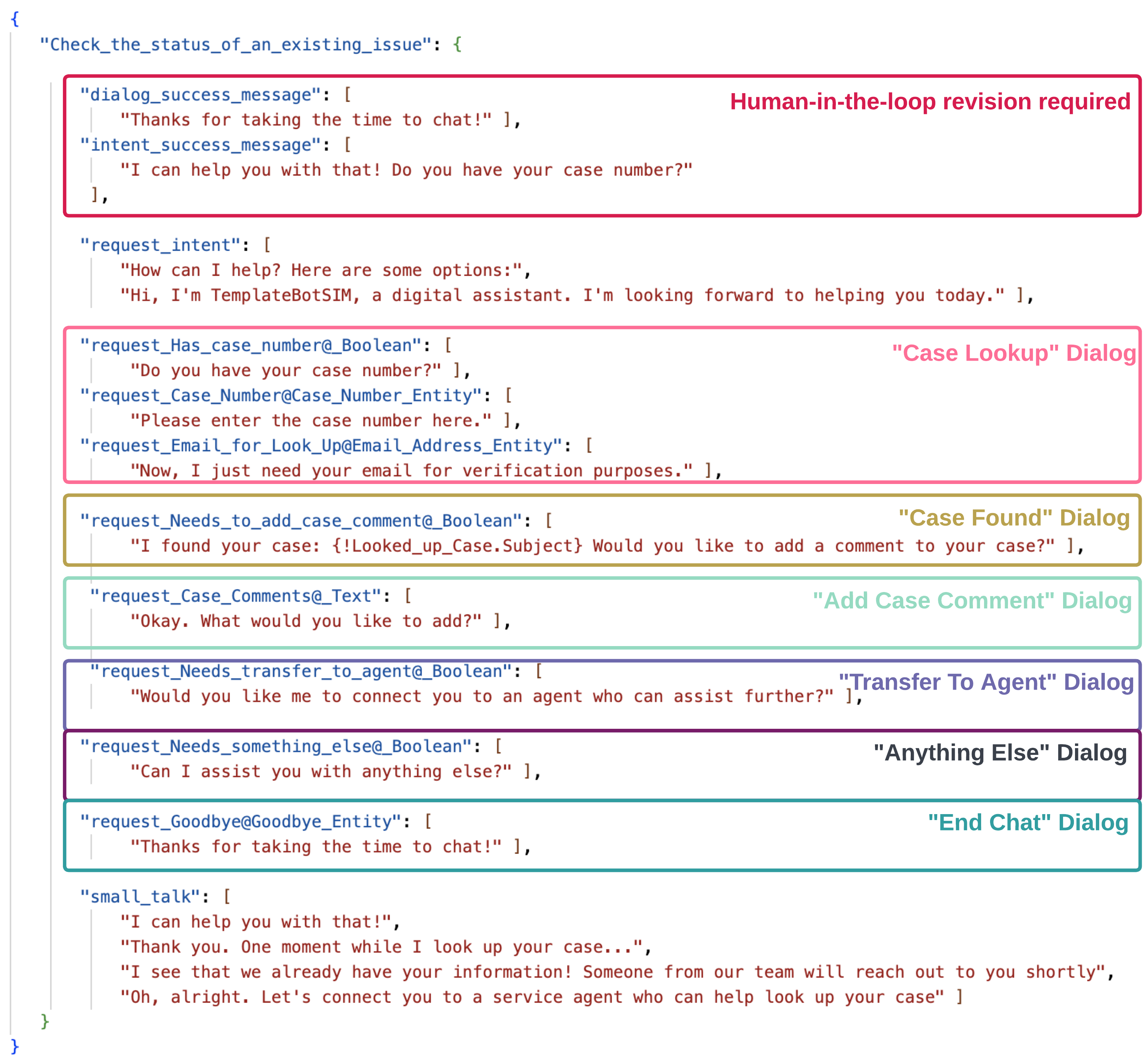} 
   \caption{Automatically generated dialog act maps for the mega dialog ``Check the status of an existing issue'' from the Salesforce Einstein BotBuilder Template Bot.}
 \label{fig:dialog_act_map}
 \vspace{-1em}
\end{figure}

The generator takes bot designs and intent utterances as input and produces the required  configuration files and dialog goals for dialog simulation.
\vspace{-0.5em}
\paragraph{Dialog act maps.} Most commercial TOD bots follow a ``rule-action-message'' design scheme and there exist clear mappings from system messages to rules/actions. For example,  the utterance ``May I get your email?'' (message) is used to ``Collect'' (action) the ``Email'' (slot) with  entity type  ``Email'' from the user. Therefore, this message can be mapped to the ``request\_Email'' dialog act by the generator parser.
As the only platform-specific component, the parser acts as an ``adaptor'' to unify bot definitions from different platforms to a common representation of dialog act maps (example in Figure~\ref{fig:dialog_act_map}) from bot messages to dialog acts. Such (local) dialog acts are automatically inferred by the parser for each dialog. Furthermore, the parser unifies the entire bot design as a graph, where individual dialogs  are vertices and their transitions are edges. Each graph node is initially associated with its ``local'' dialog act map. The dialog act map of a ``mega'' dialog containing references to other dialogs will be updated by including all the ``local'' dialog act maps of the dialog nodes along the paths  starting from the mega dialog to the terminating dialogs (\emph{e.g.,} ``End\_Chat''). The algorithm is detailed in Appendix~\ref{alg:dialog_act_map_gen}. 
The graph modeling not only enables BotSIM to  naturally support the simulation of mega-agents with multiple intents/dialogs (Figure~\ref{fig:cx-graph}), it also offers a scalable and controllable approach to exploring conversation paths for greater test coverage and potentially benefiting the conversation design as well.

The generated dialog act maps serve as the BotSIM NLU module  to map system messages to dialog acts via fuzzy matching.\footnote{\url{https://github.com/seatgeek/thefuzz}}
In particular, the two dialog acts,  ``dialog\_success\_message'' and ``intent\_success\_message''  are the  golden labels  indicating a successful dialog and a correct intent classification, respectively. They are inferred heuristically by taking the first message as  ``intent\_success\_message'' and last message as ``dialog\_success\_mesage''. 
BotSIM users are required to review or revise these two dialog acts for each evaluation dialog to confirm their correctness.
 
\paragraph{Simulation goals.} For agenda-based dialog simulation, a user goal comprises a set of dialog acts and entity slot-value pairs needed to complete the task defined by the goal. 
The dialog acts and slots are from the parsed dialog act maps and the entity values are randomly initialised according to some heuristics. As the entity values are mostly related to products/services, to better test bot NER capabilities, users can replace these random values to real ones when generating simulation goals.
Below is a snippet of a simulation goal. The goal is generated by collecting the entity-value pairs in the dialog act map and the ontology. 
The ``inform\_slots''  contains entities to be ``informed'' to the bot, whereas the ``request\_slots'' comprises entities to be ``requested'' from the bot.
    \begin{Verbatim}[fontsize=\small, commandchars=\\\{\}]
\textbf{Check_the_status_of_an_existing_issue_0}:
  \textbf{goal}: Check_the_status_of_an_existing_issue
  \textbf{inform_slots}:
    \textbf{Email_for_Look_Up}: andrews@ms-mail.com
    \textbf{Case_Number}: C379870
    \textbf{Intent}: Can I check the latest status of 
            my reported issue?
  \textbf{request_slots}:
    \textbf{Check_the_status_of_an_existing_issue}: UNK
  \textbf{...}
\end{Verbatim}
All the entity-value pairs in ``inform\_slots'' of the goals are used to test bots' NLU capabilities. The special ``intent'' slot  contains the intent queries generated by the paraphrasing models for pre-deployment testing or user-provided evaluation utterances for performance monitoring. 
\begin{table*}[h]
\centering
\begin{tabular}{cccc|ccc}
        & \multicolumn{3}{c}{WIKI-Answers} & \multicolumn{3}{c}{QQP}   \\
        \toprule
        & Target ($\uparrow$)     & Self($\downarrow$)   & iBLEU ($\uparrow$)  & Target($\uparrow$)   & Self($\downarrow$) & iBLEU($\uparrow$)  \\
        \toprule
HVQ-VAE     & 39.5     & 33.0          & 24.9    & 30.5  & 40.2      & 16.4  \\
T5-base & 33.9    & 23.9       & 23.9   & 29.1 & 35.2     & 16.3 \\
\bottomrule
\end{tabular}
\caption{Paraphrasing model comparison. BLEUs  are computed from the top one paraphrase with the reference (Target-BLEU)/input (Self-BLEU). iBLEUs are obtained by a weighted sum of target (0.8) and self BLEUs (-0.2). }
\label{paraphrasing_model_comparsion}
\vspace{-1em}
\end{table*}
\vspace{-0.5em}
\paragraph{T5 paraphrasing model.} As a core model component, we fine-tune a T5-base ~\citep{raffel2020exploring} model for paraphrasing.
To further improve the diversity, model ensemble with an off-the-shelf  Huggingface Pegasus~\citep{zhang2019pegasus} model\footnote{\url{https://huggingface.co/tuner007/pegasus\_paraphrase}} is adopted. 
{The paraphrasing models take intent utterances as input and output their top $N$ paraphrases by beam search. The paraphrases are subsequently filtered by discarding candidates with low semantic similarity scores and  edit distances. The filtered paraphrases serve as  the ``intent'' slot values of the goals as intent queries for pre-deployment testing.}
Our T5-base paraphrasing model has very competitive performance on par with the state-of-the-art HVQ-VAE model as shown in Table~\ref{paraphrasing_model_comparsion}. It is worth noting that the T5 model yields significantly lower self-BLEU scores, which means the generated paraphrases share less lexical similarities with the source sentences, a merit desirable for BotSIM in generating dialogs to cover greater breadth of language variation.  More details of the T5 model are discussed in Appendix~\ref{t5-paraphrasing}.
\vspace{-0.5em}

\subsection{Simulator}
We use a dialog-act-level ABUS rather than NUS for the following reasons. 
First, BotSIM targets commercial use cases and simulation duration and computation are crucial non-functional considerations. NUS inference usually requires GPUs, which can significantly increase the  barrier to entry and operational cost.
Second, NUS requires large amounts of annotated data to train and  are prone to overfitting. Finally, dialogue-act-level simulation is more platform{-} and task{-}agnostic.
The user simulator can be viewed as a dialog agent with its  NLU, NLG and dialog state manager.  

\vspace{-0.5em}
\paragraph{NLU.} BotSIM uses dialog act maps to map bot messages to dialog acts via fuzzy matching. 

\vspace{-0.5em}
\paragraph{NLG.} For efficient end-to-end dialog simulation, template-based NLG is adopted to convert user dialog acts to natural language responses. Given a dialog act, \emph{e.g.,} ``request\_Email'', a response is randomly chosen from a set of pre-defined templates with a ``Email'' slot, which is replaced by the value in the goal during dialog simulation. The plug-and play user response templates can be  constantly updated to include more language variation as encountered in real use cases. 

\begin{figure}[!t]
  \centering
  \includegraphics[width=7.8cm]{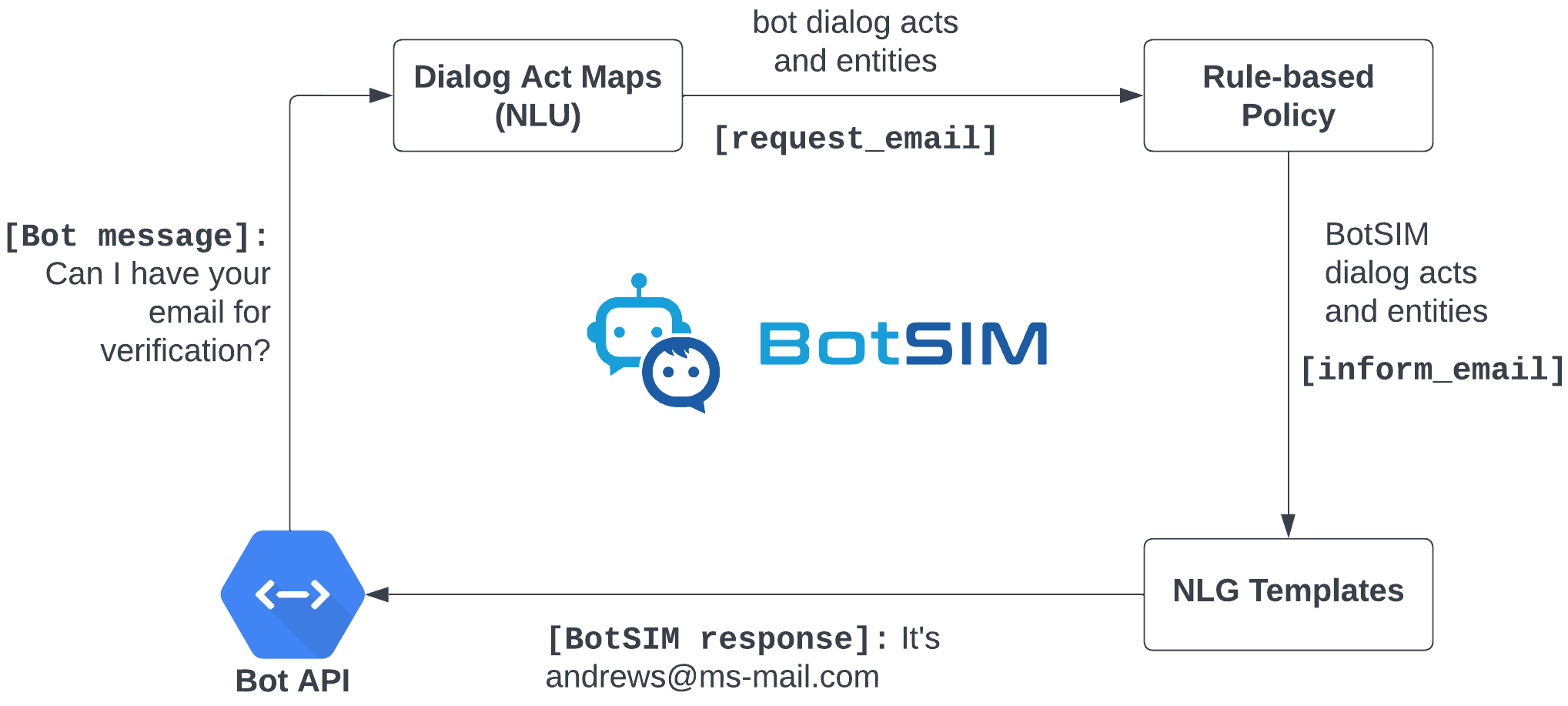} 
  \caption{A conversation turn between BotSIM and the bot during a dialog simulation.}
  \label{fig:botsim_simulation_loop}
  \vspace{-1em}
\end{figure}

\vspace{-0.5em}
\paragraph{Dialog state manager.} Rule-based dialog manager is used for its simplicity and robustness. The dialog states are maintained as a stack-like structure called agenda. During simulation, user dialog acts are popped from the agenda to respond to different system dialog acts. 
The two most important rules are for responding to ``request'' and ``inform'' dialog acts. While most of the bot behaviours/messages can be converted to these two dialog acts, BotSIM allows users to implement new rules to accommodate novel dialog acts that may only exist in their own bot designs.
Figure~\ref{fig:botsim_simulation_loop} illustrates an API-based conversation turn between BotSIM and the bot during dialog simulation: Based on the dialog acts matched by the NLU, the state manager applies the corresponding rules to generate the user dialog acts. They are then converted to natural language  responses by the NLG and sent back to the bot. The conversation ends when the task has been successfully finished or an error has been captured.

\subsection{Remediator}
The remediator generates health reports, performs analyses, and provides actionable insights to troubleshoot and improve dialog systems.
The reports are presented in a dashboard in Figure~\ref{fig:remediator-reports}, ~\ref{fig:remediator-intent}, ~\ref{fig:remediator-ner} and ~\ref{fig:remediator-analytics}. More detailed introduction is given in Appendix~\ref{rem-report-analytic}. 
\vspace{-0.5em}
 \paragraph{Bot health reports.} The bot health dashboard consists of a set of multi-level  performance reports. At the highest level, users can have a historical view of most recent simulation/test sessions (\emph{e.g.,} after each major bot update) to evaluate the impacts of bot changes from the performance trend in Figure \ref{fig:remediator-reports}(1). Users can also investigate a selected test session as in Figure \ref{fig:remediator-reports}(2). Given a test session, users can select a dialog/intent to check the detailed performance in Figure \ref{fig:remediator-reports}(3). From the detailed intent and NER plots, one can easily identify the most confusing intents and entities. 
 \vspace{-0.5em} 
 \paragraph{Actionable remediation suggestions.} 
The outputs of the Remediator comprises actionable suggestions from analysing the simulated dialogs with errors in Figure ~\ref{fig:remediator-intent}(4). 
The dashboard allows detailed investigation of all intent or NER errors together with the simulated conversation. 
The root causes of the failed conversations are identified via backtracking of the simulation agenda.
For intent models, the intent queries/paraphrases that lead to intent errors are grouped by the original intent utterances sorted by the number of errors in descending order (drop-down list of Figure ~\ref{fig:remediator-intent}(4)). Depending on the classified intent labels, the remediator would suggest some follow-up actions (Figure ~\ref{fig:remediator-intent}(5)). For example, augmenting the intent training set with the queries deemed to be out-of-domain by the current intent model, moving the intent utterance to another intent if most of  paraphrases of the former intent  utterance are classified to the latter intent. 
 \vspace{-0.5em}
 \paragraph{Conversation analytics.} Another useful component of the Remediator is the suite of conversation analytical tools to gain more insights for troubleshooting and improving their dialog systems. They include: confusion matrix analysis (Figure ~\ref{fig:remediator-analytics}(7)) for identifying confusion among intents and potential intent clusters~\citep{thoma2017analysis}, tSNE~\citep{JMLR:v9:vandermaaten08a} clustering of the sentence embeddings of intent utterances (Figure ~\ref{fig:remediator-analytics}(8)) to help evaluate the training data quality and detect intent overlaps. 
  \vspace{-0.5em}
\paragraph{Conversation graph modelling}
Powered by the  underlying conversation graph model, the conversation flow visualisation tool (Figure ~\ref{fig:remediator-analytics}(9)) helps users explore their current dialog designs. For example, users can select the ``source'' and ``target'' dialogs to investigate the generated dialog paths.  Not only is the tool valuable for comprehensive testing coverage of conversation paths, it also offers a controllable approach to troubleshooting dialog design related errors or improving the bot design.
 

\begin{table*}[!htbp]
\centering
\begin{tabular}{cccccccc}
Model  & Eval. & TA & EC & CS & CI & CO & RI   \\ 
\toprule
\multirow{2}{*}{Baseline}  
& original   &  0.92$\pm$0.03  &  0.95$\pm$0.02  &  0.89$\pm$0.03  & 0.93$\pm$0.03   & 0.94$\pm$0.02 &   0.82$\pm$0.04      \\ 
   & paraphr.  &  0.88$\pm$0.01  & 0.93$\pm$0.01   & 0.85$\pm$0.01   & 0.91$\pm$0.01  &  0.93$\pm$0.01  & 0.77 $\pm$0.02      \\
\midrule
\multirow{2}{*}{Retrained} 
   & original   & 0.92$\pm$0.03   & 0.97$\pm$0.02   &  0.93$\pm$0.03  & 0.95$\pm$0.02  & 0.96$\pm$0.02 &  0.87$\pm$0.04     \\
   & paraphr.  & 0.89$\pm$0.01   &     0.94$\pm$0.01 & 0.90$\pm$0.01 & 0.94$\pm$0.01  & 0.94$\pm$0.01 &0.80$\pm$0.02    \\
\bottomrule
\end{tabular}
\caption{Results for the Einstein Bots case study, before and after retraining the intent model with the augmented training set (F1 with 95\% confidence interval computed with 10K bootstrapped samples).}
\label{f1-comparison}
\vspace{-1em}
\end{table*}
\begin{table*}[ht]
\centering
\begin{tabular}{ccccccc}
          & CB (86)         & MP (66)        & LC (139)        & IC   (224)      & CC  (142)       & Acc   \\ 
\toprule
Baseline  & 0.84$\pm$0.06 & 0.83$\pm$0.07 & 0.88$\pm$0.04 & 0.95$\pm$0.02 & 0.96$\pm$0.02 & 0.90  \\ 
\midrule
Retrained & 0.91$\pm$0.04 & 0.89$\pm$0.06 & 0.94$\pm$0.03 & 0.95$\pm$0.02 & 0.95$\pm$0.03 & 0.92  \\ 
\bottomrule
\end{tabular}
\caption{F1 (95\% confidence interval) comparison of intent models before and after retraining for CX case study}
\label{dialogflowcx}
\vspace{-1em}
\end{table*}
\vspace{-0.5em}
\section{Case Studies}
\subsection{Salesforce Einstein Bot}
The ``Template Bot'' is the  pre-built bot of the Salesforce Einstein BotBuilder platform. It has six intents with hand-crafted training utterances. 
\vspace{-0.5em}
\paragraph{Experimental setup.}
We sample 150 utterances per intent as the training set (train-original) and use the rest for evaluation (eval-original).  The six intents are: ``Transfer to agent (TA)'', ``End chat (EC)'', ``Connect with sales (CS)'', ``Check issue status (CI)'', ``Check order status (CO)'' and ``Report an issue (RI)''.
We show how BotSIM can be used to perform data-efficient end-to-end evaluation through dialog user simulation.
To probe the baseline system, we apply the paraphrasing models to the ``train-original'' utterances to get the ``train-paraphrases'' dataset and use it as the development set. Simulation goals are created by taking the ``train-paraphrases'' as the intent queries to capture the variations in real user intent queries. The ``train-paraphrases'' goals are then used to evaluate the dialog system via dialog simulation. After simulation, the Remediator produces the performance reports and remediation suggestions.
Although the Remediator provides suggestions for remedying  both intent and NER errors, we focus on the intent model since it can be retrained (NER model has not supported retraining yet).  Another reason  is that the entity values in the goals are randomly generated and may not reflect the real-world values. 
Since the impact of the NER is removed,
the improvement of intent performance directly translates to the improvement of dialog success rate. 
It is also important to note that the suggestions are  meant to be used as guidelines rather than strictly followed. They can also be extended by users to include domain expertise. 
To validate the effectiveness of the remediation suggestions, we augment the recommended misclassified paraphrases  to the ``train-original'' set to form the ``train-augmented'' set and retrain the intent model. We then compare the performance before and after retraining on the goals created from the ``evaluation-original''.
\vspace{-0.8em}
\paragraph{Results and analytics.}
We observe consistent improvements for all intents on the human-written ``eval-original'' set   after model retraining. More challenging intents (lower F1s), \emph{e.g.,} ``RI'' and ``CS'', saw larger performance gains compared to the easier ones such as ``EC'' (higher  F1s). This demonstrates the efficacy of BotSIM and is likely due to more paraphrases being selected for retraining the model on the more challenging intents. 
We applied the paraphrasing models to the ``eval-original'' set to get the ``eval-paraphrases'' set to further increase the test coverage. 
In Table~\ref{coverage}, the second row (\xmark) shows the number of misclassified ``eval-original'' utterances. Out of the remaining correctly classified ``eval-original'' utterances in the first row(\cmark), substantially larger number of them have at least one of their paraphrases in ``eval-paraphrases''  wrongly classified by the same intent model. This 
 indicates that the diversity introduced by the paraphrasing models potentially expands the test coverage by a large margin. 
 \vspace{-0.5em}
\subsection{Google DialogFlow CX}
We use the pre-built financial service mega-agent for the flow-based evaluations. 
Even for a single flow in Figure~\ref{fig:cx-graph}, it is non-trivial to 
manually design conversations to cover all  paths. Through BotSIM's conversation graph modeling, the flow-based conversations can be simulated  by generating goals consisting of the dialog acts of all dialog nodes along a traversal path. 
On top of these flow-based dialog paths, paraphrases of the intent utterances can be used as the intent queries to probe the NLU performance via dialog simulation.
To simulate pre-deployment testing, we choose five flows and split the intent utterances into train and evaluation sets. The intent F1 scores are given in Table~\ref{dialogflowcx}. The flows are ``Check Balance (CB)'', ``Make Payment (MP)'', ``Lost Card'', ``Investigate Charges(IC)'', ``Compare Cards (CC)''.
Since the financial bot has only $\sim$30 utterances for training each intent, to obtain a more reliable test set, we use the ``eval-paraphrases'' set together with the ``eval-original''. The total number of evaluation intent queries are inside the parentheses of the 
 Table~\ref{dialogflowcx} header. 
Similar to the previous study,  retrained intent model outperforms the baseline in terms of both F1 and accuracy (Table~\ref{dialogflowcx}), especially for the challenging flows such as ``CB'', ``MP''.
\vspace{-0.5em}
\section{Conclusion}
We presented BotSIM,  a modular end-to-end bot simulation toolkit for multi-intent dialog generation and evaluation of commercial TOD systems via agenda-based dialog user simulation. 
Our case studies show that BotSIM can save substantial manual effort in bot evaluation, troubleshooting and improvement. BotSIM can be easily extended to support new platforms by implementing a set of well-defined parser functions to convert bot messages to dialog acts. We are in the midst of open-sourcing the codes including the Web App. We also plan to support more platforms as future work.
\section{Limitations}
For efficiency reasons, BotSIM adopts a template-based NLG model for converting user dialog acts to natural languages. Although the template-NLG is more controllable and flexible compared to the model-based NLG, they may lack naturalness. One possible future improvement includes a combination of template-based NLG and the model-based NLG. For example, we can train a model-based NLG to generate templates~\citep{DBLP:journals/corr/abs-1808-10122} for BotSIM's response templates. In this way, both efficiency and naturalness can be achieved.
\section{Broader Impact}
The pretrained language{-}model based paraphrasers (T5-base and Pegasus) used in this study are pretrained and finetuned with large scale of text corpora scraped from the web, which may contain biases. These biases may even be propagated to the generated paraphrases, causing harm to the subject of these stereotypes. Although the paraphrasing models are only applied to generate the testing intent queries, BotSIM users are advised to take into consideration these ethical issues and may wish to manually inspect or otherwise filter the generated paraphrases. 
It is also noteworthy that to prevent any data privacy leakage, the information produced in the simulation (the entity values in the BotSIM ontology) is randomly generated, and therefore fake. This includes the email addresses, names. 
Currently, BotSIM is trained and evaluated utilizing English text. We leave multi-lingual bot simulation capability as one of our future works.
\bibliography{anthology,custom}
\bibliographystyle{acl_natbib}

\appendix

\section{Regression test cases for Botium}
\label{test-case}
Below is a Botium regression testing case (script):
\begin{Verbatim}[fontsize=\small]
T01_Check_Card_Balance
#me
What is the due amount for my card?
#bot
Can I have the last 4 digit card number?
INTENT balance_enquiry
#me
5789.
#bot
You have $50.8 due for this card.
INTENT balance_enquiry
ENTITY_VALUES 5789
\end{Verbatim}

It is important to note that the Botium ``end-to-end'' testing is different from BotSIM as it aims to ensure the bot interface and operation work across different devices without regression errors. In other words, they are still for detecting regressions rather performance evaluation. For BotSIM's end-to-end evaluation, we refer to the setup that BotSIM takes in bot designs and automatically 1)  parses  bot designs, 2) generates test cases, 3) performs large-scale end-to-end dialog simulations, 4) analyzes outputs and provides actionable remediation suggestions for troubleshooting and improvement.

\begin{algorithm*}
\caption{\texttt{Dialog act maps inference from bot designs (MetaData/API)}}\label{alg:dialog_act_map_gen}
\begin{algorithmic}
 \State{\texttt{local\_dialog\_act\_maps = \{\}}}
\For{\texttt{dialog $\in$ all\_dialogs}}
    \State{\texttt{local\_dialog\_act\_maps [dialog] = \{\}}}
    \For {\texttt{message $\in$ all\_messages }}
        \State \texttt{dialog\_act = infer\_dialog\_act\_from\_message(message)}
        \If {\texttt{dialog\_act not in local\_dialog\_act\_maps [dialog]}}
            \State{\texttt{local\_dialog\_act\_maps [dialog][dialog\_act] = []}}
        \EndIf
         \State \texttt{local\_dialog\_act\_maps [dialog][dialog\_act].append(message)}
    \EndFor
\EndFor

\State{\texttt{global\_dialog\_act\_maps = \{\}}}
\For{\texttt{dialog $\in$ all\_non\_end\_dialogs}}
\State{\texttt{global\_dialog\_act\_maps[dialog] = local\_dialog\_act\_maps[dialog]}}
\For{\texttt{end\_dialog $\in$ all\_end\_dialogs}}
\For{ \texttt{node $\in$ conv\_graph.simple\_paths(dialog, end\_dialog)}}
    \State{\texttt{global\_dialog\_act\_maps[dialog].update(local\_dialog\_act\_maps[node])}}
\EndFor
\EndFor
\EndFor
\end{algorithmic}
\end{algorithm*}

\section{T5-base paraphrasing model}
\label{t5-paraphrasing}
Pytorch~\citep{NEURIPS2019_9015} is used to fine-tune the T5-base model with the adam optimizer. The effective batch size is 128. The learning rate is 1e-4 and no warm-up is applied. The maximum sequence length is 128. Four NVIDIA A100 GPUs are used. After each epoch, we compute the iBLEU scores on the development set according to ~\cite{hosking-etal-2022-hierarchical,sun-zhou-2012-joint} to decide whether to keep the current checkpoint or not. The best model was obtained after 88 epochs.
\subsection{Paraphrase datasets}
We use the datasets released in the HVQ-VAE paper~\cite{hosking-etal-2022-hierarchical} for finetuning the paraphrasing model. The datasets are ``Wiki-Answers'', ``QQP'' and ``MSCOCO''. Instead of training a separate model for each dataset as in~\cite{hosking-etal-2022-hierarchical}, we fintune a single paraphrasing model from the pooled dataset of the three. Additionally, we also curate our own set of paraphrasing datasets and use them together with the other three datasets. The dataset information is given in Table~\ref{t5-corpura}.
Note some of the datasets are not initially designed for paraphrasing tasks. Therefore, they may also contain noisy or trivial labels for paraphrasing. Therefore, a filtering process is applied to select the final training sentence pairs.
The filtering process applies thresholds on semantic and lexical scores to strike a good balance between lexical variation and semantic similarity between the paraphrasing  sentence pairs.  In particular, we use sentence transformer~\citep{reimers2019sentencebert} score to measure the semantic similarity and FuzzyWuzzy ratio (based on  Levenshtein distance~\footnote{\url{https://pypi.org/project/python-Levenshtein/}}) for lexical diversity. Sentence pairs with low semantic similarities (noisy labels) or low Levenshtein distances (trivial labels with little lexical variation) are discarded.
\begin{table*}[h]
\centering
\begin{tabular}{ccccc}
          & Task          & Sent-Transformer score & FuzzRatio & No. Final Pairs  \\ 
\toprule
SNLI      & NLI           & {[}0.70, 0.99]           & -           & 13,635                 \\
MNLI      & NLI           & {[}0.80, 0.99]           & -           & 32.398                 \\
PAWS-Wiki & Paraphrasing  & -                       & -           & 19,004                 \\
tapaco-en & ~Paraphrasing & ~[0.50, 0.99]            &  70         & 14,735                 \\
\midrule
WIKI-Answers & Paraphrasing & - & - & 79,6679 \\
QQP & Paraphrasing & - & - & 16,3621 \\
MSCOCO & Paraphrasing & - & - & 47,3210 \\
\bottomrule
\end{tabular}
\caption{Datasets for T5-base paraphrasing model finetuning. ``-'' means no filtering applied.}
\label{t5-corpura}
\end{table*}
\begin{table*}[h]
\centering
\begin{tabular}{cccc|ccc}
        & \multicolumn{3}{c}{WIKI-Answers} & \multicolumn{3}{c}{QQP}   \\
        \toprule
        & Target ($\uparrow$)     & Self($\downarrow$)   & iBLEU ($\uparrow$)  & Target($\uparrow$)   & Self($\downarrow$) & iBLEU($\uparrow$)  \\
        \toprule
Pegasus &  31.4        &     55.3        &   14.0      & 23.8 & 46.0     & 9.9  \\
HVQ-VAE     & 39.5     & 33.0          & 24.9    & 30.5  & 40.2      & 16.4  \\
T5-base-Wiki & 42.7 & 42.7 & \textbf{25.6} & 14.9 & 20.0 & {7.9*} \\
T5-base-QQP & 32.3 & 46.8 & 16.5* & 31.9 & 42.7 & \textbf{17.0}\\
T5-base-Multi-Task & 33.9    & 23.9       & 23.9   & 29.1 & 35.2     & 16.3 \\
\bottomrule
\end{tabular}
\caption{Performance benchmarking of different paraphrasing models. The BLEU scores are computed from the top one paraphrase candidate  with respect to the reference (Target-BLEU) or the input (Self-BLEU) sentence. The iBLEU scores are calculated using $\texttt{Target-BLEU}*0.8 - \texttt{Self-BLEU}*0.2$. Numbers with asterisk* denote the ``zero-shot'' performance.}
\label{detailed_paraphrasing_model_comparsion}
\end{table*}
We also performed a benchmark of iBLEU scores in Table~\ref{paraphrasing_model_comparsion}. 
The paraphrases are generated via beam search with the same beam size of 10. The results of HVQ-VAE are taken from the original paper. The discrepancy of the QQP results from the paper is due to some train/eval data overlap we found in their original setup. We contacted the authors and they provided the updated QQP results and the fixed datasets. We thus used the bug-fixed version for finetuning and evaluating our T5-base model. 
The off-the-shelf Pegasus model has the largest model size but performed the worst compared to the other two models across all scores. Since the author did not reveal anything about the model, we cannot finetune it with the same datasets as the T5-base. 
Using the same setup as HVQ-VAE, we finetuned a T5-base-Single-Task model for each task and it consistently outperformed the HVQ-VAE model on the task that it was trained on. On the contrary, The single-task models performed significantly worse on the task they were not trained on (see the numbers with *), indicating the negative impacts of domain mismatch. In addition, it is impractical to finetune a new model for each new dataset or task. Therefore, we pooled all datasets together and finetuned a single model ``T5-base-Multi-Task''. Although the multi-task model performs slightly worse than the single-task ones on each individual task, the overall performance is still on par with the state-of-the-art HVQ-VAE model, especially on the QQP task. Therefore, we choose the ``T5-base-Multi-Task'' as the BotSIM paraphrasing model.

\begin{figure}[!t]
  \centering
  \includegraphics[width=8cm]{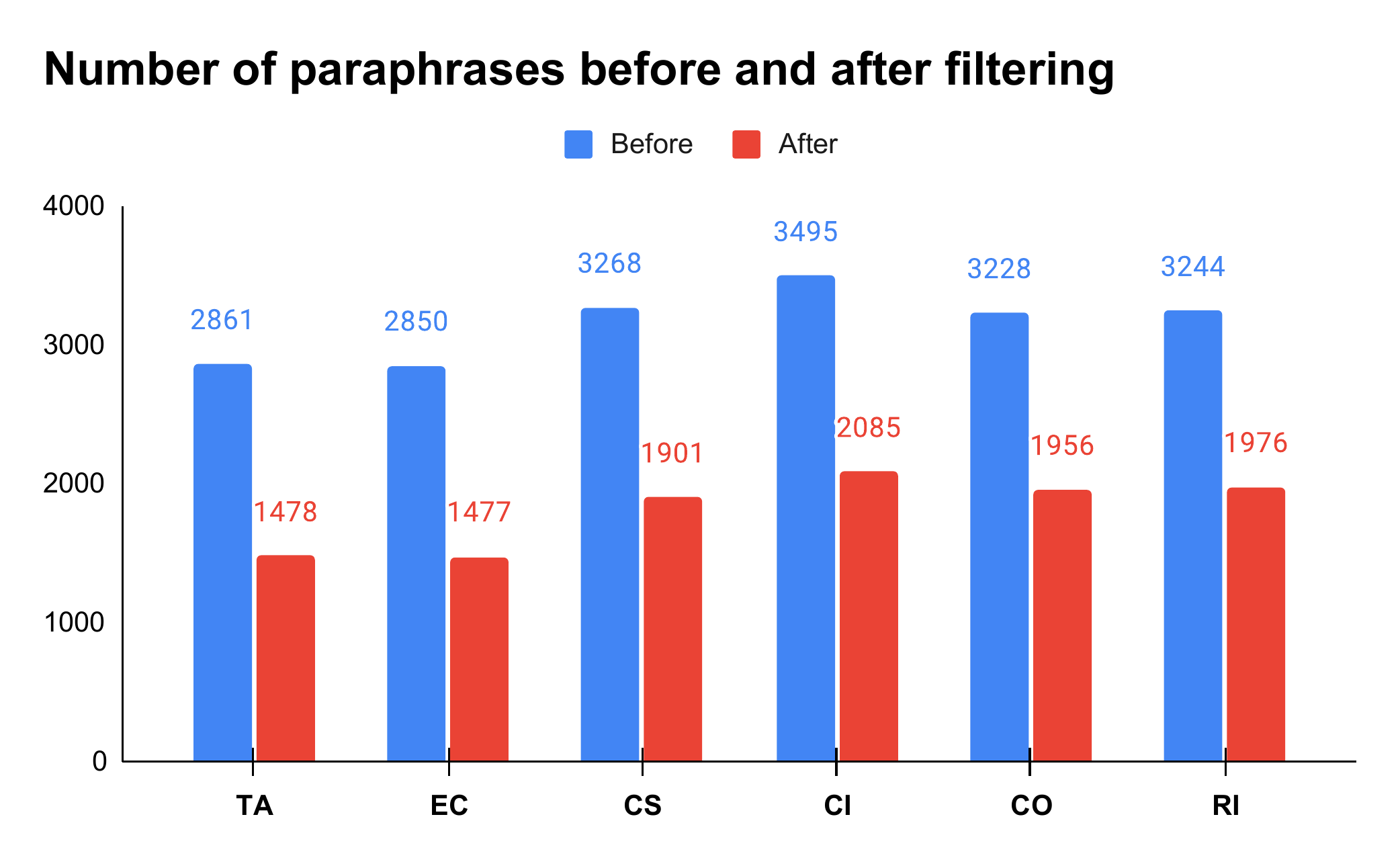} 
  \caption{Number of paraphrase candidates before and after semantic and lexical filtering for ``train-paraphrases''}
  \label{fig:filter}
\end{figure}

We apply the same filtering principle for the training data preparation to the generated paraphrases to keep the ones with high semantic  and low lexical similarities.  In Figure~\ref{fig:filter}, we show the number of candidates before and after filtering when generating the ``train-paraphrases'' set.
For simple intents like ``TA'' and ``EC'', almost half of the original  candidates are discarded. More challenging intents have more surviving paraphrases due to larger variation in their training utterances. The filtered candidates are then used as the intent queries for creating the simulation goals.

\begin{table*}[t]
\centering
\begin{tabular}{cccccccc}
Dataset                & Intent enquiries & TA  & EC  & CS  & CI  & CO  & RI   \\ 
\toprule
\multirow{2}{*}{Train} & train-original             & 150 & 150 & 150 & 150 & 150 & 150  \\
                       & train-augmented  & 255  &  184       &  212  &  268   & 215    &  294       \\ 
\midrule
Dev                    & train-paraphrases & 1465   &  1467      &  1754  &  1989  & 1895   &  1786      \\ 
\midrule
\multirow{2}{*}{Eval}  & eval-original   &   182 & 145   &  183  &  222  & 205   & 178      \\
                       & eval-paraphrases   &  1190  & 933   &  1648  & 2172   &    1936 &   1795\\
\bottomrule
\end{tabular}
\caption{Dataset information for the Einstein Template Bot case study. }
\label{dataset}
\end{table*}

\begin{table}[ht]
\centering
\begin{tabular}{ccccccc}
 eval-original & TA & EC & CS & CI & CO & RI  \\ 
\toprule
 prediction \cmark         & 9  & 17 & 27 & 33 & 34 & 61  \\ 
 prediction \xmark              & 9  & 7  & 16 & 19 & 8  & 26  \\ 
\bottomrule
\end{tabular}
\caption{Test coverage expansion via paraphrasing}
\label{coverage}
\end{table}
\subsection{Investigation into misclassified paraphrases}
From Table~\ref{coverage}, we can see the paraphrasing models help increase the testing coverage as some correctly classified original intent queries (prediction \cmark) have misclassified paraphrase intent queries (prediction \xmark). 
Below we show two successfully classified original utterances with their wrongly classified paraphrases of the ``Report an issue (RI)'' intent. 
\begin{lstlisting}[label={paraphrase},language=Python]
{
   My order was damaged in shipping: [
    Why did my order get damaged during shipping?,
    Why was my order damaged when it was shipped?,
    Shipping damaged my order.,
    The order was damaged when it was shipped.
  ],
  The bottle of conditioner was open when it arrived i need a replacement: [
    I need a new bottle of conditioner because the one that arrived was open.,
    I need a replacement for the opened bottle of conditioner.,
    I need a replacement for the open bottle of conditioner.,
    I need a new bottle of conditioner because the one I received was open.
  ]
 }
\end{lstlisting}
As suggested by the Remediator, users can select some of the high-quality  paraphrases to augment the original intent training set to refine the intent models. They can also filter from the evaluation (manual-crafted utterances or product chat logs) paraphrases to create a larger evaluation set for performance monitoring. This saves substantial human efforts in creating or annotating dialog testing data.

\section{Remediator reports and analytical tools}
\label{rem-report-analytic}
\begin{figure*}[t]
  \centering
\includegraphics[width=17cm]{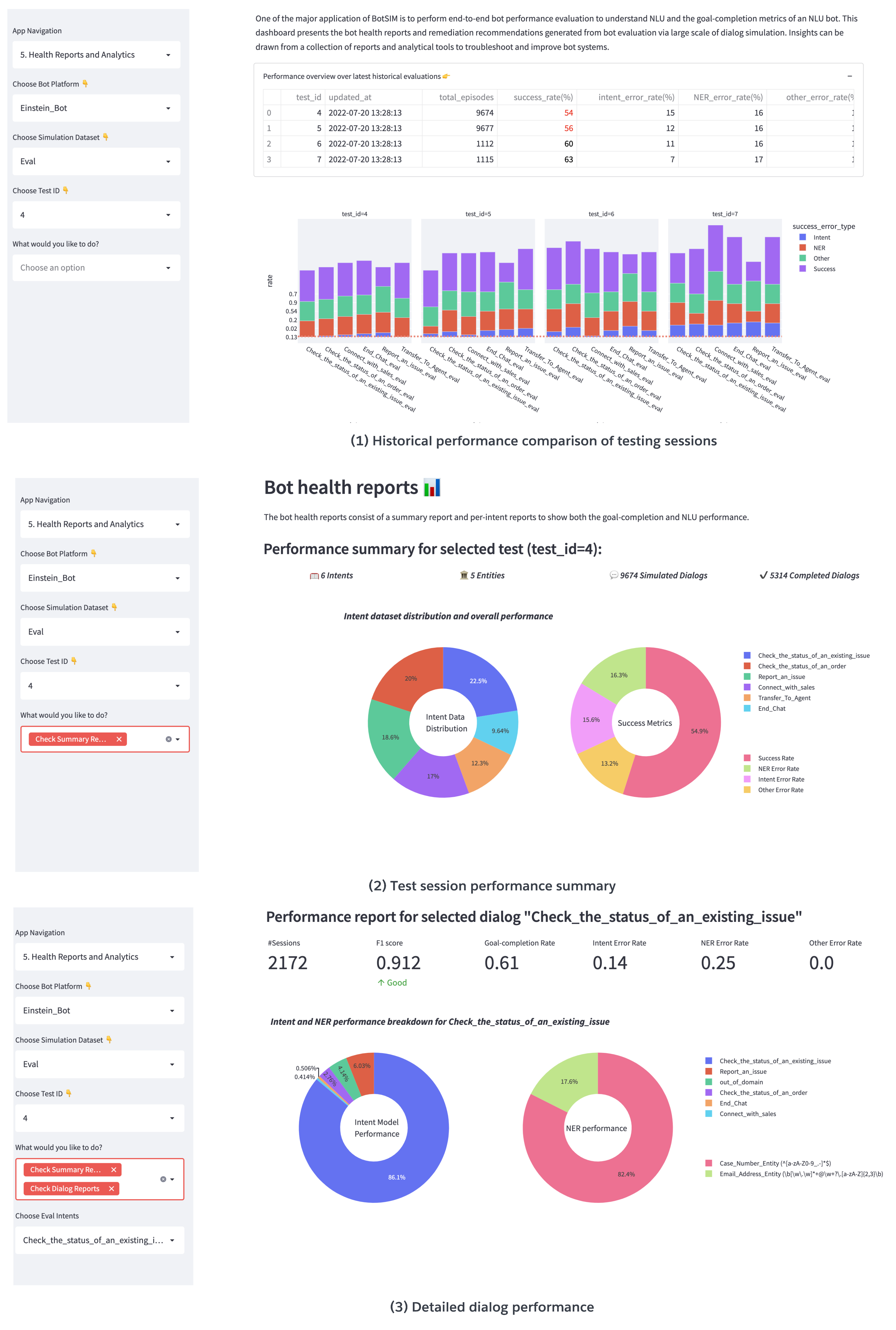} 
  \caption{Remediator dashboard: bot health reports. }
  \label{fig:remediator-reports}
\end{figure*}

\begin{figure*}[t]
  \centering
\includegraphics[width=17cm]{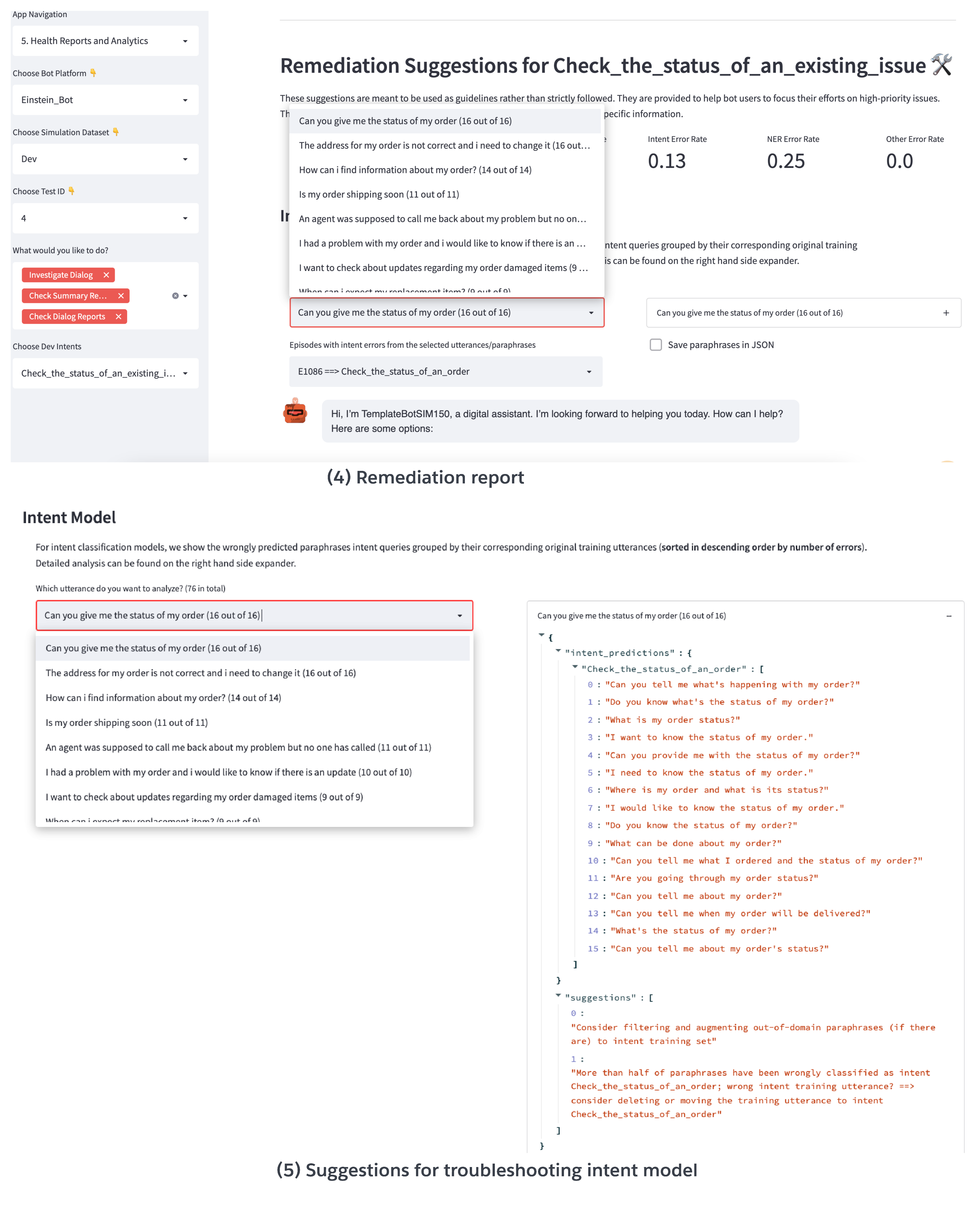} 
  \caption{Remediator dashboard: intent model remediation suggestions. }
  \label{fig:remediator-intent}
\end{figure*}

\begin{figure*}[t]
  \centering
\includegraphics[width=15cm]{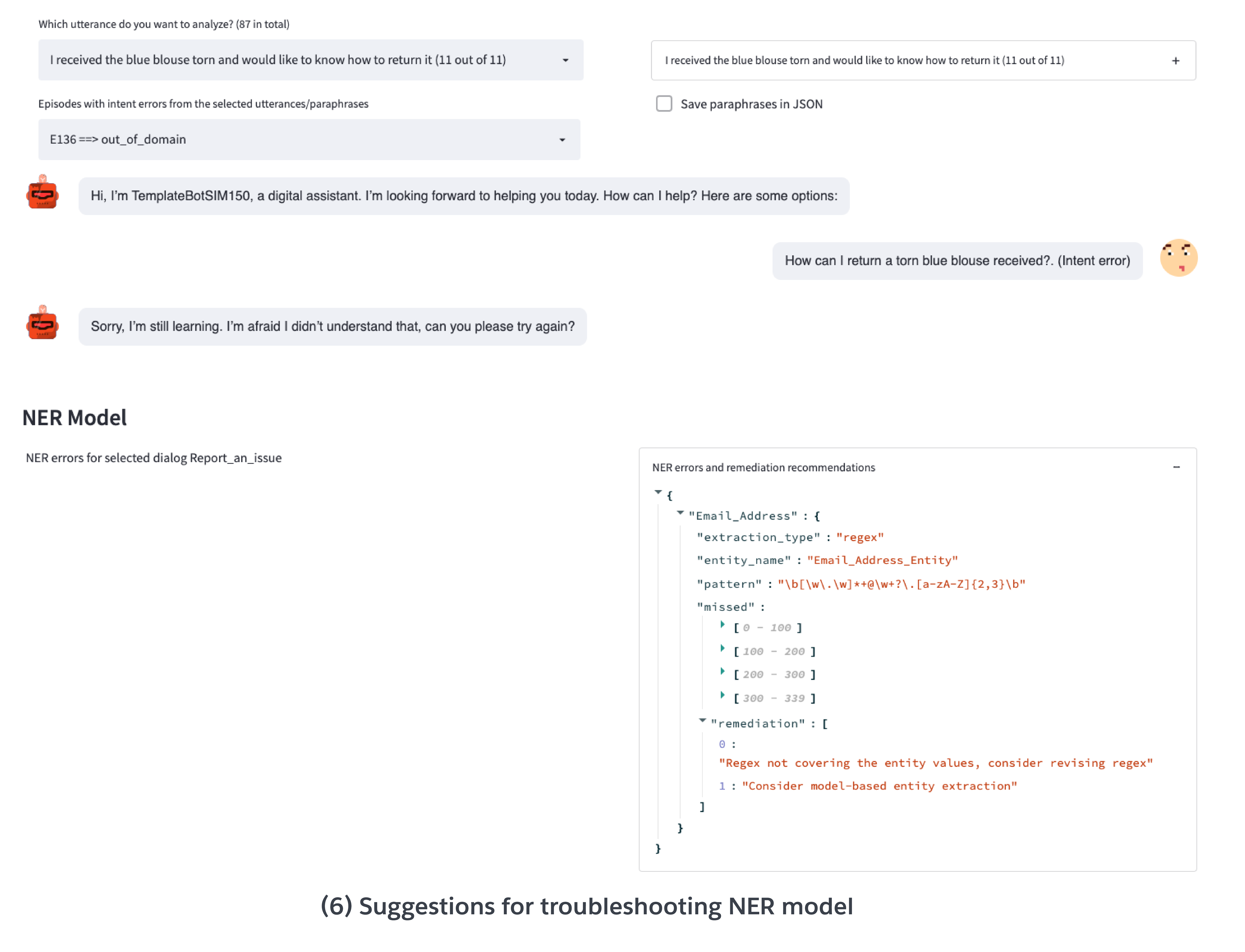} 
  \caption{Remediator dashboard: NER remediation suggestions. }
  \label{fig:remediator-ner}
\end{figure*}

\begin{figure*}[t]
  \centering
\includegraphics[width=17cm]{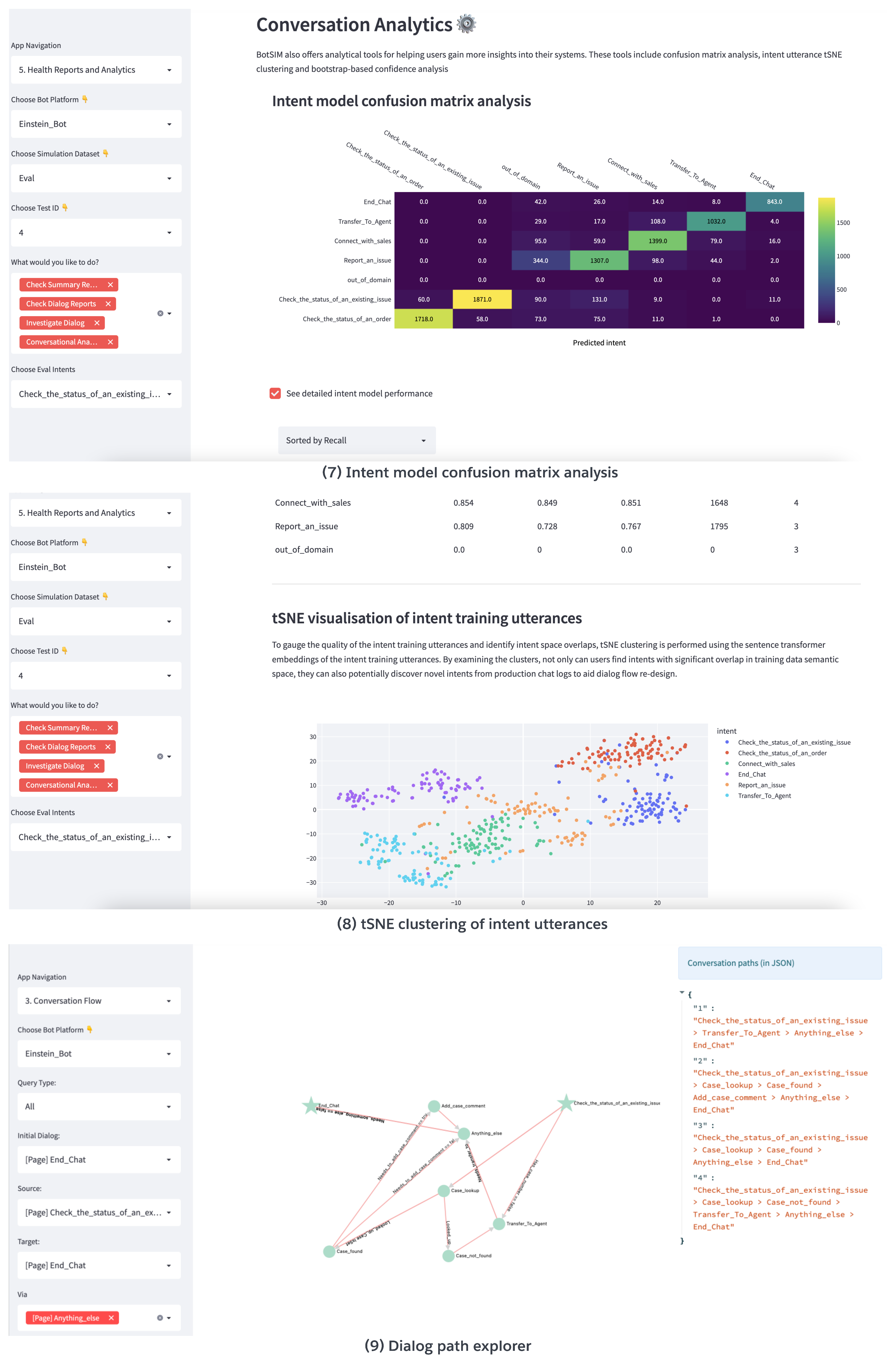} 
  \caption{Remediator dashboard: Conversational analytical tools. }
  \label{fig:remediator-analytics}
\end{figure*}
The Remdediator outputs are detailed in the bot health report dashboard shown in Figure~\ref{fig:remediator-reports}, ~\ref{fig:remediator-intent}, ~\ref{fig:remediator-ner} and ~\ref{fig:remediator-analytics}. 
The left panel gives users options to navigate through the dashboard. For example, they can select different  bot platforms, datasets, test ids and intents. 
The bot health reports in Figure~\ref{fig:remediator-reports} offer a multi-granularity view of simulation performance. At the highest level is the historical comparison of most recent testing sessions (Figure~\ref{fig:remediator-reports}(1)). For example, a testing session may be needed after each major bot update. From the historical performance comparison, users can see how certain changes impact the overall bot performance and decide whether to keep or revert the update. 

Given the historical performance, users may be interested in further investigating a particular testing session. They can do so by selecting one from the drop-list of all testing sessions and enable the ``Check Summary Report'' option in the multi-selection box. The resulting overall bot health report for the selected test session is shown in Figure~\ref{fig:remediator-reports}(2). It summarizes the simulation information including number of intents, entities and simulation episodes. The two doughnut charts depict the dataset distribution and the overall success metrics.

To investigate the detailed performance report of  each individual intent (Figure~\ref{fig:remediator-reports}(3)), users can navigate to ``Check Dialog Report'' and select a dialog from the drop-list. The detailed dialog report presents the intent and NER performance.  One can quickly identify the most confusing intents or entities and focus their efforts to investigate and resolve the confusions.

To help troubleshoot the identified errors, users can select ``Investigate Dialog'' to see the remediation suggestions. Figure~\ref{fig:remediator-intent}(5) shows the misclassified intent query paraphrases and their corresponding original utterances, grouped by the wrongly predicted intent labels. Given the prediction results, suggestions are provided for possible further actions. For the given example, all paraphrases of the utterance ``Can you give me the status of my order''  have been classified as the ``check order'' intent, indicating an annotation error of the original utterance. Therefore, this utterance should be moved from the ``check issue'' intent to the ``check order'' intent.

To gain more insights into their bot systems, users can harness the conversation analytical tools for better comprehension of the simulation results. 
To understand more about the intent classifier, confusion matrix analysis is applied to the intent predictions of the simulated conversations(in Figure~\ref{fig:remediator-analytics}(7)). A detailed and sortable intent performance can be displayed by checking the checkbox, allowing users to quickly identify the worst performing intents in terms of recall, precision or F1 rates. This helps them  plan and allocate resources to improve the poor-performing intents.

To gauge the quality of the intent training utterances and identify intent overlaps,  tSNE clustering is performed based on the sentence transformer embeddings of the intent training utterances. By examining the clusters, not only can  users find intents with significant overlaps in the training data semantic space, they can also potentially discover novel intents from production chat logs to aid dialog flow re-design. The dashboard can be easily extended to to support more analytical tasks.

\section{Streamlit web app}
Finally, we give some brief discussions of the Streamlit Web App. The motivation is to offer  BotSIM not just as a framework for developers but also as an easy-to-use  app to end users such as bot admins without diving into  technical details.
 The app can be deployed as a docker container or to the Heroku platform.
We use Streamlit as the front-end and Flask as the backend. A set of API functions are designed to communicate with BotSIM.
For multi-platform support,  keeping track of the simulation status and historical performance, a SQL-based database is used. 
 BotSIM supports two types of databases including Sqlite3 and Postgres. 
 To support Heroku deployment, particularly its  ephemeral file system,  cloud storage such as AWS S3 is used to store the simulation logs and results.

\end{document}